\documentclass{article}
\usepackage{cite}
\usepackage{spconf,amsmath,epsfig}
\usepackage{algorithmic}
\usepackage{algorithm}
\usepackage{multirow}
\usepackage{enumitem}

\let\OLDthebibliography\thebibliography
\renewcommand\thebibliography[1]{
  \OLDthebibliography{#1}
  \setlength{\parskip}{0pt}
  \setlength{\itemsep}{0pt plus 0.3ex}
}

\begin{document}\sloppy
\topmargin=0mm

\def\x{{\mathbf x}}
\def\L{{\cal L}}

\title{Revisiting Multi-Granularity Representation via Group Contrastive  Learning for Unsupervised Vehicle Re-identification}

\twoauthors{Zhigang Chang
                      \thanks{This work was supported by the NSFC62071292, U21B2013.}}
{Shanghai Jiao Tong University \\
                        changzig@sjtu.edu.cn}
{Shibao Zheng}
{Shanghai Jiao Tong University \\
                        sbzh@sjtu.edu.cn}                      



\maketitle

\begin{abstract}
Vehicle re-identification (Vehicle ReID) aims at retrieving vehicle images across disjoint surveillance camera views. The majority of vehicle ReID research is heavily reliant upon supervisory labels from specific human-collected datasets for training. When applied to the large-scale real-world scenario, these models will experience dreadful performance declines
due to the notable domain discrepancy between the source dataset and the target. To address this challenge, in this paper, we propose an unsupervised vehicle ReID framework (MGR-GCL). It integrates a multi-granularity CNN representation for learning discriminative transferable features and a contrastive learning module responsible for efficient domain adaptation in the unlabeled target domain. Specifically, after training the proposed Multi-Granularity Representation (MGR) on the labeled source dataset, we propose a group contrastive learning module (GCL) to generate pseudo labels for the target dataset, facilitating the domain adaptation process. We conducted extensive experiments and the results demonstrated our superiority against existing state-of-the-art methods.
\end{abstract}

\section{Introduction}
\label{sec:introduction}
Vehicle re-identification (Vehicle Re-ID) seeks to associate vehicle images captured by disparate surveillance cameras. As one of the most practical and deployable computer vision tasks ~\cite{CycleGAN2017, chang2020weighted, yang2020large, liu2021graph}, vehicle Re-ID has witnessed notable breakthroughs with adequate labeling deep learning theory and the advent of tremendous convolutional neural network-based models \cite{liu2016deepa,sun2018beyond,zhao2021phd}. Nevertheless, when applied to the real-world wide-scale scenario, these models will inevitably experience fateful performance deterioration due to the substantial domain discrepancy. Domain adaptation, which seeks to bridge the domain gap between the source and target datasets, has been grabbing growing research attention.

Recently, a few domain adaptation methods for Re-ID tasks have been proposed. Specifically, it mainly includes three lines of research. The first one seeks to translate the source images into target images with Cycle GAN\cite{CycleGAN2017} based models and train the model with the generated pseudo labeled target images to bridge the domain gap directly. The second line of work is dedicated to learning domain-agnostic representations between the source and target datasets\cite{zhang2018domain,xu2020adversarial}, through which the model learns to extract domain-invariant features. Though achieving pronounced performance improvements, these models still fall far behind their supervised counterparts since they rely entirely on the supervisory signals from the source dataset, leaving the characteristic embedded in the unlabeled target dataset unexploited. In light of this limitation, \cite{bashir2019vr,song2020unsupervised} proposed to adopt an unsupervised clustering method to generate pseudo labels for the target dataset and, then, train the model with the pseudo labels in an iterative fashion to further boost the performance on the target dataset. However, these methods rely on primary convolution neural networks, which take as input the vehicle images and output a holistic feature representation, thus, ignoring fine-grained transferable features and leading to compromised domain adaptation performance.

In light of these limitations, we propose a novel domain adaptation framework, named MGR-GCL, to tackle the unsupervised vehicle re-identification problem. Concretely, inspired by \cite{sun2018beyond}, which learns fine-grained features through vertically partitioning the middle-level feature maps, we designed a novel dual-directional fine-grained part-level feature learning network specifically for the vehicle re-identification scenario. The basic intuition is that, unlike person images which exhibit more discriminative features in the vertical directions, the main discriminative and transferable characteristics are embodied in the horizontal direction for vehicle images. Another intuition is that learning fine-grained part-level features is beneficial for the transfer learning task since fine-grained features incorporate less domain-specific information and more domain-common features than their holistic counterparts.

\begin{figure*}[tbp]
\vskip -0.6cm
\includegraphics[width=\textwidth]{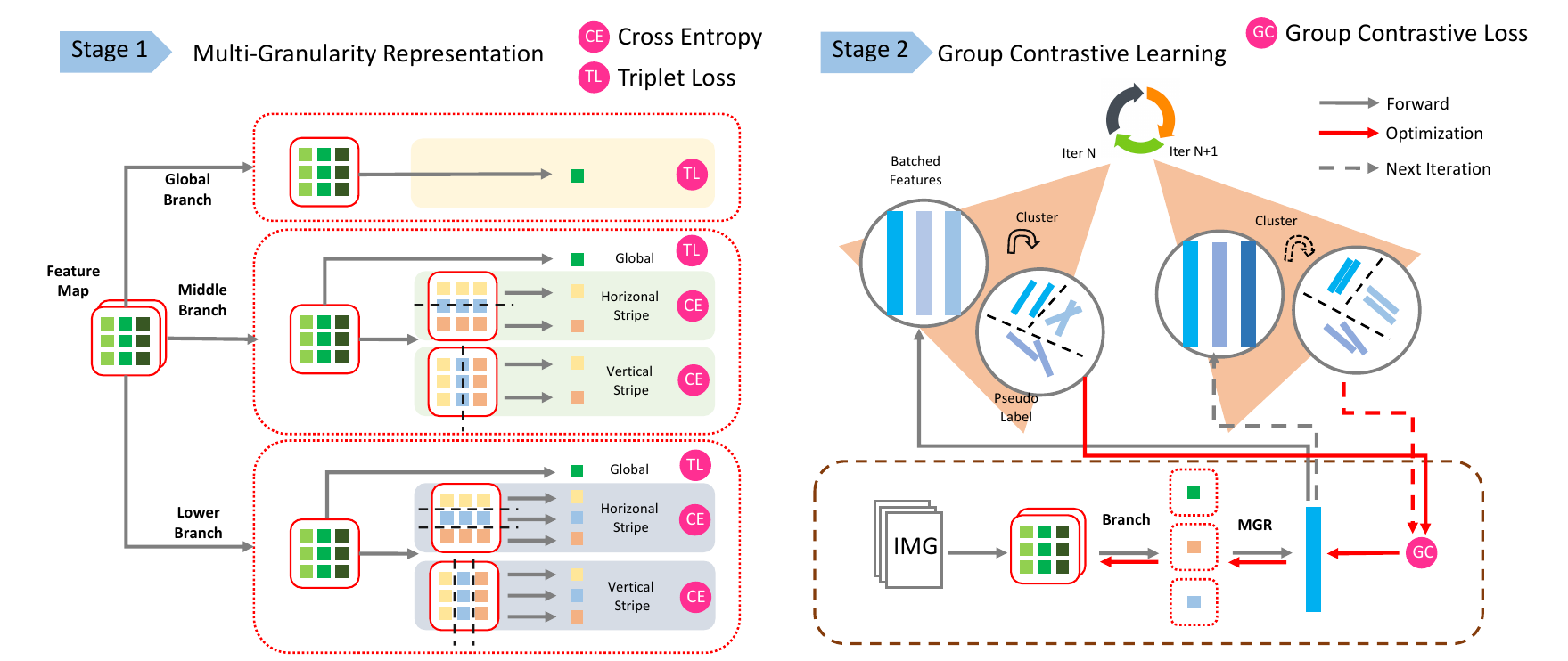}
\caption{The framework of MGR-GCL. It basically contains two stages. Stage 1 is a supervised multi-granularity representation learning process in the source domain. Stage 2 illustrates the unsupervised domain adaptation stage by group contrastive learning.}
\label{fig1}
\vskip -0.4cm
\end{figure*}

Furthermore, inspired by recent advancements in contrastive learning \cite{he2020momentum,ge2020self,dai2022cluster,chang2019distribution}, to efficiently mine the data characteristics in the unlabeled target domain, we propose a group contrastive learning module (GCL) to generate pseudo target labels and train the model with the generated labels in the target domain to further boost adaptation. Specifically, after training multi-granularity representation (MGR) on the source dataset, we concatenate the fine-grained part features into a large feature representation and utilize the DBSCAN clustering method to generate group labels for the target dataset. In the later stage, we compute a contrastive loss in a group level to update the network. The pseudo group generation and contrastive learning are conducted in an iterative fashion until the model converges. The detailed framework of our proposed method is elucidated in Fig.~\ref{fig1}. To sum up, our major contributions are as follows:

\begin{itemize}
\item We propose a novel domain adaptation framework, named \textbf{MGR-GCL}, for unsupervised vehicle re-identification.

\item A multi-granularity neural network architecture is proposed to learn fine-grained transferable features from two directions. Then, a novel group contrastive learning module (GCL) is introduced to facilitate adaptation by leveraging target dataset characteristics.

\item Comprehensive experiments on VeRi, VehicleID and VehicleX, have been conducted. We show that our method outperforms the state-of-the-arts in all the settings.

\end{itemize}

\vskip -0.5cm
\section{Related Works}
\subsection{Contrastive Learning}
Contrastive learning~\cite{he2020momentum,ge2020self,dai2022cluster} has demonstrated significant success in unsupervised representation learning. It functions by denoting the anchor and its randomly-augmented counterpart as similar and the rest images in the batch as dissimilar. Then, a contrastive loss such as InfoNCE is adopted to pull close similar images in the representation and push away dissimilar ones. It generally requires large batch sizes to achieve satisfactory results, introducing significant GPU memory consumption. \cite{he2020momentum} further proposes a momentum-updated memory module to decouple the training with the batch sizes. Inspired by \cite{dai2022cluster}, different from these works adopting an instance contrastive learning paradigm, we propose a novel iterative group contrastive learning by performing contrastive learning at a group level and in an iterative fashion for domain adaptation.

\subsection{Unsupervised Domain Adaptation}
Unsupervised Domain Adaptation. In light of the beforementioned shortcomings, unsupervised domain adaptation methods seeking to bridge the domain gap have been investigated. Recently, some works in this community strive to transfer source data into target data in the raw pixel space to alleviate the domain discrepancy directly~\cite{CycleGAN2017}. Other works focus on learning domain-invariant features by aligning the feature distribution of the source with the target~\cite{zhang2018domain,xu2020adversarial}. For instance, \cite{ganin2015unsupervised} proposed an adversarial learning-based method to efficiently fuse the source and target distribution to learn domain-invariant features. Despite the notable progress achieved in this domain, traditional unsupervised domain adaptation methods do not apply to the vehicle reidentification scenario since they are based on the assumption that class labels are identical between the source and the target datasets, which is not the case for ReID problems.

\vspace{-0.3cm}
\section{Proposed Method}
\subsection{Problem Definition}
For unsupervised domain adaptation in the vehicle Re-ID scenario, we have a well-labeled source dataset $D_{S}=\left \{ X_{S},Y_{S} \right \}$ and a totally-unlabeled target dataset $D_{T} = \left \{ X_{T} \right \}$. Concretely, $X_{S}=\left [ x_1, x_2, \dots ,x_{N_s} \right ]$ is the source vehicle images where $N_s$ is the total number of the images of the dataset. $Y_{S}=\left [ y_1, y_2, \dots ,y_{N_s} \right ]$ is the corresponding label set for the source dataset. For the target dataset, there are $N_T$ vehicle images without any labels. The goal of unsupervised domain adaptation is to enhance the vehicle Re-ID performance on the target dataset $D_T$ only using the supervisory information from the source dataset $D_S$.

\subsection{Fully Supervised Multi-Granularity Representation Learning}
In this paper, we propose a novel transfer learning backbone architecture to learn Multi-Granularity Representation of vehicles, which is also a discriminative transferable feature between the source and target datasets. As demonstrated in stage 1 of Fig~\ref{fig1}, we devise a dual-directional part-based architecture. In particular, for extracting basic feature maps, we adopt the ResNet-50 layer structure before $res\_conv4\_1$. Subsequently, we divide the feature map after $res\_conv4\_1$ into three independent branches. For the upper branch, stride-2 convolutional layer in $res\_conv\_5$ is employed for downsampling, which is, then, followed by a global max-pooling (GMP) operation and squeeze operation to obtain a 2048-dim feature vector. We name this branch as Global Branch and the corresponding feature vector as $v^g$.

For the Middle and Lower branches, we adopt a similar architecture as the Global Branch without down-sampling to preserve more salient local features. For both two branches, we obtain two global feature 2048-dim feature vectors, namely $v^m$ and $v^l$. Then, for each branch, we adopt a dual-directional partitioning strategy.

\noindent \textbf{Vertical Part Partitioning.} For the Middle Branch, we perform vertical partitioning on the output feature map to obtain two independent feature stripes. Then, separate $1\times1$ convolution layers with batch normalization and a ReLU activation layer are employed to reduce the 2048-dim feature vector into 256-dim feature representations, which are named $p^{M}_{v,1}$ and $p^{M}_{v,2}$ respectively. In a same fashion, we partition the Lower Branch into three stripes and, consequently, obtain three feature vectors, $p^{L}_{v,1}$, $p^{L}_{v,2}$ and $p^{L}_{v,3}$.

\noindent \textbf{Horizontal Part Partitioning.} We propose to further learn fine-grained features in the horizontal direction by partitioning the output feature maps from the middle and lower branch into two and three strips horizontally. Accordingly, we obtain corresponding feature vectors $p^{M}_{h,1}$ and $p^{M}_{h,2}$ for the Middle Branch and $p^{L}_{h,1}$, $p^{L}_{h,2}$ and $p^{L}_{h,3}$ for the Lower Branch.

\noindent \textbf{Supervised Source Training.} Fully supervised learning scheme is conducted on the source dataset $D_S$ with the proposed MGR model. Precisely, we utilize triplet loss and softmax cross-entropy loss on the global feature set $S_{g} = \left \{ v^{g},v^{m},v^{l}  \right \} $ and local feature set $S_{l} = \left \{ p^{M}_{v,1}, p^{M}_{v,2},p^{L}_{v,1},p^{L}_{v,2},p^{L}_{v,3},p^{M}_{h,1}, p^{M}_{h,2},p^{L}_{h,1},p^{L}_{h,2},p^{L}_{h,3}\right \} $, respectively. We use batch hard triplet loss for hard negative sample mining, paired with a random identity sampler where each batch samples $P$ identities and $K$ instances for each identity. The triplet loss conducted on global feature set $S_g$ are denoted as follows:
\begin{equation}
\begin{aligned}
    L_{\mathrm{Triplet}} & =\frac{1}{\left | S_g \right | } \sum_{f\in S_g}\sum_{i=1}^{P}\sum_{a=1}^{K}[ \alpha + \overbrace{\max \limits_{p=1\dots K }\left \| f_{a}^{(i)}-f_{p}^{(i)}  \right \|_{2}}^{\mathrm{hardest\ positive} } \\ & - \underbrace{ \min \limits_{\substack{n=1 \dots K \\ j=1 \dots P} }\left \| f_{a}^{(i)} - f_{n}^{(j)} \right \|_{2} }_{\mathrm{hardest\ negative} } ]   
\end{aligned}
\label{eq1}
\end{equation}

where $f_a$,$f_p$,$f_n$ are the features for the anchor, positive and negative vehicle images in the batch used for triplet loss calculation, respectively. $\alpha$ is the margin. Aside from exploiting triplet loss, we also employ softmax cross-entropy loss in local feature set $S_{l}$, regarding each vehicle identity as a specific class, which is noted as Identity Loss $L_{\mathrm{ID}}$. The overall supervised source data training loss function is the combination of Triplet Loss and Identity Loss, denoted as follows:
\begin{equation}
    L_{\mathrm{Supervised}} = L_{\mathrm{Triplet}} + L_{\mathrm{ID}}
\label{eq2}
\end{equation}

\subsection{Unsupervised Pseudo Label Generation}
\label{sec:label}
After sufficiently trained on the source dataset $D_S$, when directly applied to the target dataset $D_T$, DPN will still experience
performance drops owing to the sizable domain discrepancy. Motivated by \cite{dai2022cluster}, we adopt DBSCAN (Density-Based Spatial Clustering of Applications with Noise), an efficient clustering method, to generate clusters in the target dataset $D_T$ since it does not require the predefined number of clustering centroids. Unlike previous methods, which utilize
the holistic feature representation to perform clustering, we instead propose to synthesize a robust mega feature vector $f = v^{g}\oplus v^{l}\oplus \dots \oplus p^{L}_{h,3}$ by concatenating the feature representations in feature sets $S_g$ and $S_l$. Then, for every unlabeled image $x^{i}_{t}$ in target dataset $D_T$, we can obtain a  $f^{i}_{t}$, leading to a set of features as $S_f= \left \{ f^{1}_{t},f^{2}_{t},\dots,f^{N_t}_{t} \right \}$.

In the later stage, we employ the unsupervised clustering method DBSCAN on the feature set $S_f$ to generate a series of groups. Then, every unlabeled image $x^{i}_{t}$ is assigned a pseudo label $y^{i}_{t}$ according to the group it resides in. So a newly-labeled target dataset $D^{0}_{TN}$ is established, denoted as below:
\begin{equation}
    D^{0}_{TN} =\left \{ \left ( x^{i}_{t}, y^{i}_{t}\right ) \mid x^{i}_{t}\in D_T \right \} 
\label{eq3}
\end{equation}

\subsection{Group Contrastive Feature Learning}
Finally, to in the new target dataset $D^{0}_{TN}$, we propose a group contrastive learning paradigm given the pseudo group labels learned in Sec.~\ref{sec:label}. Similar
to \cite{he2020momentum}, we adopt a memory $C = \left \{ c_1, \dots ,c_{N_c} \right \}$  of size $N_C$ to store the representative group features where $N_C$ is the number of groups in Sec.~\ref{sec:label}. The memory is initialized with the mean feature vector for each group as:
\begin{equation}
    c_k = \frac{1}{\left | \mathcal{H}_{k}  \right | } \sum_{f_i\in \mathcal{H}_{k}} f_i
\label{eq4}
\end{equation}
where $\mathcal{H}_{k}$ denotes the feature set for $k$-th group, $f$ is a feature vector for a specific group and $\left | \cdot  \right | $ represents the number of instances for each group.

\noindent \textbf{Group Contrastive Loss.} We randomly sample $P$ groups
and for each group we randomly sample $K$ images, resulting in a batch of  query images in total. Subsequently, for each query feature $f_q$ of the image in the batch, we compute the group contrastive loss with each representative feature in the memory as:
\begin{equation}
  L_{\mathrm{GCL}} = -\log \frac{\exp (f_q \cdot c^{+})/\tau }{ {\textstyle \sum_{i=0}^{K}} \exp (f_q \cdot c_{i})/\tau} 
\label{eq5}
\end{equation}
where $c^{+}$ is a positive group feature to the query feature $f_q$ in the memory $C$ and $\tau$ is the temperature hyper-parameter.

\noindent \textbf{Group Memory Updating.} After each step of minimizing
$L_{\mathrm{GCL}}$, we update the group memory $C$ as suggested in :
\begin{equation}
\begin{aligned}
  c_k &\gets \frac{1}{\left | \mathcal{H}_k  \right | } \sum_{\substack{f_i\in\mathcal{H}_k\\f_i\in\mathcal{Q} } }[mf_i+(1-m)f_q^i] \\
  &=m \frac{1}{\left | \mathcal{H}_k  \right | } \sum_{\substack{f_i\in\mathcal{H}_k\\f_i\in\mathcal{Q} } }f_i +(1-m) \frac{1}{\left | \mathcal{H}_k  \right | }\sum_{\substack{f_i\in\mathcal{H}_k\\f_i\in\mathcal{Q} } }f_{q}^{i} \\
  &=mc_k + (1-m)q_k
\end{aligned}
\label{eq6}
\end{equation}
where $\mathcal{Q}$ is the query features in the current batch and $m$ is
a momentum parameter. In this equation, the $k$-th group representative
feature is updated with the mean of current query feature $f_q$ belonging to the $k$-th group, which is denoted as $q_k$.

\noindent \textbf{Learning Detailed Procedure.} Our proposed method involves the process of the supervised source training, memory dictionary initialization and memory updating. The learning details are presented in 
Algorithm~\ref{algo1}.

\vspace{-0.5cm}
\begin{algorithm}[h]
    \caption{Overall learning pipeline}
    \begin{algorithmic}[1] 
        \REQUIRE {Labeled Source Data $D_S$}
        \REQUIRE {Unlabeled Target Data $D_T$}
        \REQUIRE {Our proposed Network with ImageNet-pretrained ResNet-50 weights $f_{\theta}$} 
        \REQUIRE {Hyperparameters for loss functions and Memory Updating}
        \FOR{i \textbf{in} $\left [ 1, \mathrm{supervised \ num\_epochs} \right ]$ }{
            \STATE {Extract feature vectors $S_g$, $S_l$ from $D_S$ by Network $f_{\theta}$ }
            \STATE {Compute the supervised loss by Eq.~\ref{eq2} }
            \STATE {Use SGD optimizer updating the weights $f_{\theta}$}
        }
        \ENDFOR
        \FOR{j \textbf{in} $\left [ 1, \mathrm{unsupervised \ num\_epochs} \right ]$}{
            \STATE {Extract feature vectors $S_f$ from $D_T$ by Network $f_{\theta}$}
            \STATE {Cluster $S_f$ into $N_c$ groups with \textbf{DBSCAN}}
            \STATE {Initialize memory dictionary with Eq.~\ref{eq4}}
        }
            \FOR{k \textbf{in} $\left [ 1, \mathrm{num\_iterations} \right ]$}{
                \STATE {Sample P $\times$ K query images from $D_T$}
                \STATE {Compute $L_{GCL}$ loss with Eq.~\ref{eq5}}
                \STATE {Update cluster features with Eq.~\ref{eq6}}
            }
            \ENDFOR
        \ENDFOR
    \end{algorithmic}
\label{algo1}
\end{algorithm}

\vspace{-0.5cm}
\begin{table*}[th]
\begin{center}
\caption{
Performance of state-of-the-art methods on \textbf{VeRi-776} to \textbf{VehicleID}}
\label{table2}
\resizebox{18cm}{!}{
\begin{tabular}{l|c|ccc|ccc|ccc|ccc}
\hline
\multirow{2}{*}{Methods} &  \multirow{2}{*}{source}&  \multicolumn{3}{c|}{Test size = 800} & \multicolumn{3}{c|}{Test size = 1600} &  \multicolumn{3}{c|}{Test size = 2400} & \multicolumn{3}{c}{Test size = 3200}\\
\cline{3-14} & & mAP & rank1 & rank5  & mAP & rank1 & rank5 & mAP & rank1 & rank5 & mAP & rank1 & rank5\\
\hline \hline
FACT &VeRi-776 & - & 49.53 & 67.96 & - & 44.63 & 64.19 & - & 39.91 & 60.49 & - & - & - \\
Mixed Diff+CCL &VeRi-776 & - & 49.00 & 73.50 & - & 42.80 & 66.80 & - & 38.20 & 61.60 & - & - & - \\
PUL &VeRi-776 & 43.90 & 40.03 & 56.03 & 37.68 & 33.83 & 49.72 & 34.71 & 30.90 & 47.18 & 32.44 & 28.86 & 43.41 \\
CycleGAN &VeRi-776 & 42.32 & 37.29 & 58.56 & 34.92 & 30.00 & 49.96 & 31.89 & 27.15 & 46.52 & 29.17 & 24.83 & 42.17 \\
DT baseline &VeRi-776 & 40.58 & 35.48 & 57.26 & 33.59 & 28.86 & 48.34 & 30.50 & 26.08 & 44.02 & 27.90 & 23.85 & 39.76 \\
\hline \hline
PAL &VeRi-776 & 53.50 & 50.25 & 64.91 & 48.05 & 44.25 & 60.95 & 45.14 & 41.08 & 59.12 & 42.13 & 38.19 & 55.32 \\
PLM &VeRi-776 & 54.85 &51.23 & 67.11 &49.41 &45.40 &63.37 &46.00 &41.73 &60.94 &43.46 &39.25 &57.99 \\
MGR-GCL &VeRi-776 & \textbf{55.24} & \textbf{52.38} & \textbf{75.29} & \textbf{50.56} &\textbf{45.88} &\textbf{67.65} &\textbf{47.59} & \textbf{42.83} & \textbf{64.36} & \textbf{44.36} & \textbf{40.07} & \textbf{59.82} \\

\hline \hline
\end{tabular}}
\end{center}
\vspace{-0.3cm}
\end{table*}

\vspace{-0.2cm}
\begin{table}[ht]
\caption{Performance of Methods on \textbf{VehicleID} to \textbf{VeRi-776}}
\begin{center}
\label{table3}
\resizebox{8cm}{!}{
\begin{tabular}{r|c|ccc}
\hline
\multirow{2}{*}{Methods} & \multirow{2}{*}{source}& \multicolumn{3}{c}{VehicleID$\rightarrow$VeRi-776 } \\
\cline{3-5} & & mAP & rank1 & rank5 \\
 \hline\hline
FACT & VehicleID &18.75 & 52.21 & 72.88 \\
PUL & VehicleID &17.06 &55.24 & 66.27 \\
CycleGAN & VehicleID &21.82 &55.42 & 67.34 \\
\hline
MMT & VehicleID &35.3 &74.6 & 82.6 \\
SPCL & VehicleID &38.9 &\textbf{80.4} & 86.8 \\
PAL  & VehicleID &42.04 &66.17 & 79.91 \\
PLM  &VehicleID &47.37 &77.59 &87.00 \\
\hline
MGR-GCL & VehicleID &\textbf{48.73} &79.29 &\textbf{87.95}  \\

\hline\hline

\end{tabular}}
\end{center}
\vspace{-0.5cm}
\end{table}

\begin{table}[ht]
\caption{Performance of Methods on \textbf{VehicleX} to \textbf{VeRi-776}}
\begin{center}
\label{table4}
\resizebox{8cm}{!}{
\begin{tabular}{r|c|cccc}
\hline
\multirow{2}{*}{Methods} &  \multirow{2}{*}{source}&  \multicolumn{4}{c}{VehicleX$\rightarrow$VeRi-776 } \\
\cline{3-6} & & mAP & rank1 & rank5 & rank10 \\
 \hline\hline
MMT & VehicleX &35.6 &76.0 & 83.1 & 87.4 \\
SPCL & VehicleX &38.3 &82.1 &87.8 &90.2 \\
UCF  & VehicleX &40.6 &84.4 &88.4 &91.5 \\
\hline
MGR-GCL & VehicleX &\textbf{43.7} &\textbf{86.1} &\textbf{90.8} &\textbf{93.0}  \\

\hline\hline
\end{tabular}}
\end{center}
\end{table}

\vspace{0.8cm}
\section{Experiments}
\subsection{Datasets and Evaluation Metrics}
We employ 3 typical datasets for Vehicle Re-ID to conduct our experiments: \textbf{VeRi-776}, \textbf{VehicleID} and \textbf{VehicleX}.

\noindent \textbf{Veri-776} is a large-scale urban surveillance vehicle
dataset for reID, which contains over 50,000 images of 776 vehicles, where 37,781 images of 576 vehicles are employed as the training set, while 11,579 images of 200 vehicles are employed as a test set. A subset of 1,678 images in the test set generates the query set. \textbf{VehicleID} is a surveillance dataset from the real-world scenario, which contains 221,763 images corresponding to 26,267 vehicles in total. From the original testing data, four subsets, which contain 800, 1,600, 2,400 and 3,200 vehicles, are extracted for vehicle search for multi-scales. \textbf{Vehicle} is a large-scale synthetic dataset. Created in Unity, it contains 1362 vehicles of various 3D models with fully editable attributes.
\noindent Two widely-adopted evaluation metrics: cumulative matching characteristics (\textbf{CMC}) and mean average precision (\textbf{mAP}) are adopted for accurately and efficiently gauging the performance of our model.

\vspace{-0.2cm}
\subsection{Implementation Details}
The backbone is adapted from ResNet-50. The input images are resized to 384 $\times$ 384 and normalized. Then, standardized data augmentation techniques like random cropping, and random erasing are adopted. The margin parameter for triplet loss is set to 0.5. During training. SGD optimizer with an initial learning rate of $1e-2$ and warm-up learning scheduler are utilized to optimize the parameters.
\vspace{-0.2cm}
\subsection{Comparison with the State-of-the-arts}
We compare our proposed model with other state-of-the-art methods on three transfer scenarios VeRi$\rightarrow$VehicleID, VehicleID$\rightarrow$VeRi and VehicleX$\rightarrow$VeRi in Table. ~\ref{table2},~\ref{table3}, and~\ref{table4}. The first two settings are designed to demonstrate the performance of domain adaptation between real scenario datasets, while the third setting is designed to explore the performance of synthetic data to real scenario data. In general, in all three transfer scenarios, our method MGR-GCL significantly outperforms existing state-of-the-art approaches.

Specifically, we include supervised methods FACT~\cite{liu2016deepb}, Mixed Diff+CCL~\cite{liu2016deepa}, unsupervised Person Re-ID method PUL~\cite{fan2018unsupervised}, and a typical style domain adaptation model CycleGAN~\cite{CycleGAN2017}. DT baseline denotes applying the well-trained model on the source to the target domain by Zheng’s Method~\cite{zheng2017discriminatively}. We further incorporate the state-of-the-art method for detailed analysis: PAL~\cite{peng2020unsupervised}, MMT~\cite{ge2020mutual}, SPCL~\cite{ge2020self} and PLM~\cite{wang2022progressive}. Specifically, for adaptation from VeRi to VehicleID, as presented in Tab.~\ref{table2}, we achieve 55.24\%, 50.56\%, 47.59\%, and 44.36\% of mAP in 4 test modes, surpassing the second best PLM by over 1\%. From VehicleID to VeRi, PLM and PAL achieve notable performances with 47.37\% and 42.04\% mAP, respectively. Our method is a winner by obtaining 79.29\% and 48.\% of Rank@1 and mAP. As for cases from VehicleX to Veri, our model's performance is also firmly at the top of the benchmark, demonstrating excellent transfer capability from synthetic data to real scenarios.
\subsection{Ablation Studies}
Variants containing some of the components are qualitatively compared with the full model in this section, namely, OneIter, MGR, MGR w/o H, and MGR w/o HV. The meanings are as follows:\\
\noindent \textbf{OneIter:}  the full model after one iteration training on the target;  \textbf{MGR:} the direct-transfer MGR model without GCL; \textbf{MGR w/o H:} the MGR model without horizontal part partitioning; \textbf{MGR w/o HV:} the MGR model without horizontal and vertical part partitioning. The final results are presented in Tab~\ref{table5}. The final integrated full model consistently surpasses all the variants under all the metrics and testing scenarios.

\begin{table}[ht]
\vspace{-0.3cm}
\caption{Performance of variants containing some of the components}
\begin{center}
\label{table5}
\resizebox{8cm}{!}{
\begin{tabular}{l|cc|cc}
\hline
\multirow{2}{*}{Conponents} &   \multicolumn{2}{c|}{VehicleID$\rightarrow$VeRi} &\multicolumn{2}{c}{VehicleX$\rightarrow$VeRi} \\
\cline{2-5} & mAP & rank1 & mAP & rank1 \\
 \hline\hline
\textbf{OneIter}  &35.88 &66.21 &36.2  &69.0  \\
\textbf{MGR w/o H}  &31.24 &58.43 &30.3 &60.4 \\
\textbf{MGR w/o HV}  &29.47 & 56.45 &27.5 &53.9 \\
\textbf{MGR}  & 32.66 & 62.18 &35.9 &66.5 \\

\hline
\textbf{Full Model} &\textbf{48.73}  &\textbf{79.29} &\textbf{43.7} &\textbf{86.1}  \\

\hline\hline

\end{tabular}}
\end{center}
\end{table}
\vspace{-0.5cm}
\noindent \textbf{The effectiveness of group contrastive learning module.}
We include solo \textbf{MGR}, \textbf{OneIter} and \textbf{Full model} for comparison to validate the iterative adaptation learning scheme. The full model exceeds its original MGR by over 16\% in mAP and 17\% in Rank@1 and mAP from VehicleID to VeRi. The full model also exhibits consistent improvements with over 13\% in Rank@1 and near 13\% in mAP. These combined evidenced the effectiveness of the group contrastive feature learning module and the necessity of our iterative adaptation learning scheme.

\noindent \textbf{The effectiveness of multi-granularity representation.}
To efficiently validate the necessity of the proposed MGR with a dual-directional partitioning scheme, we compared the performance of MGR with MGR w/o H and MGR w/o HV. Without the horizontal partitioning strategy, the model experiences noticeable performance drops in both mAP and Rank@1, 1.42\% and 3.75\%, respectively. When further deprived of the vertical partitioning, the decline is even more severe, 3.19\%, 5.73\% for mAP and Rank@1, respectively. All these combined illustrated the efficacy of the proposed multi-granularity representation in learning more transferable features.
\vspace{-0.3cm}
\section{conclusion}
\vspace{-0.3cm}
We propose a novel unsupervised vehicle Re-ID framework, dubbed MGR-GCL. It combines a multi-granularity representation (MGR) with an iterative adaptation scheme that generates pseudo labels by clustering target images into different groups and then performing group contrastive learning (GCL) in an iterative fashion. We performed extensive experiments on three transfer benchmarks and the results evidenced the validity of our proposed framework.

\vspace{-0.3cm}

\bibliographystyle{IEEEbib}
\bibliography{icme2023template}

\end{document}